\documentclass[runningheads]{llncs}
\usepackage{graphicx}
\usepackage[utf8]{inputenc}
\usepackage[strings]{underscore}
\usepackage{amssymb}
\usepackage{amsmath}

\begin{document}
	\title{Deconvolving convolution neural network for cell detection}
	\author{Shan E Ahmed Raza\inst{1,2} \and
		Khalid AbdulJabbar\inst{1,2} \and
		Mariam Jamal-Hanjani\inst{3} \and
		Selvaraju Veeriah \inst{3} \and
		John Le Quesne \inst{4,5} \and
		Charles Swanton\inst{3,6,7} \and
		Yinyin Yuan\inst{1,2}}
	\authorrunning{S.E.A. Raza et al.}
	%
	\institute{Division of Molecular Pathology, The Institute of Cancer Research, London, UK. \and
		Centre for Evolution and Cancer, The Institute of Cancer Research, London, UK. \and
		Cancer Research	UK Lung Cancer Centre of Excellence, University	 College London
		Cancer	Institute,	London, UK. \and
		Leicester Cancer Research Centre, University of Leicester, Leicester, UK. \and
		MRC Toxicology Unit, University of Cambridge,  Leicester, UK. \and
		The	Francis	Crick	Institute,	London, UK. \and
		University	College	London	Hospitals	NHS	Foundation	Trust,	London, UK.}
	\maketitle              
	\begin{abstract}
		Automatic cell detection in histology images is a challenging task due to varying size, shape and features of cells and stain variations across a large cohort. Conventional deep learning methods regress the probability of each pixel belonging to the centre of a cell followed by detection of local maxima. We present deconvolution as an alternate approach to local maxima detection. The ground truth points are convolved with a mapping filter to generate artifical labels. A convolutional neural network (CNN) is modified to convolve it's output with the same mapping filter and is trained for the mapped labels. Output of the trained CNN is then deconvolved to generate points as cell detection. We compare our method with state-of-the-art deep learning approaches where the results show that the proposed approach detects cells with comparatively high precision and F1-score.
		
	\end{abstract}
	\section{Introduction}
	Cell detection/segmentation is an essential part of automated image analysis pipelines for studying the tumour microenvironment at cell level \cite{kainz2015you,Sirinukunwattana2016,Ronneberger2015,Raza2018}. This is a challenging problem due to varying size, shape and morphology of cells across the tumour landscape. Cell detection is often preferred over segmentation as it is easier to detect the cells with weak boundaries or if the nuclei are clumped together making it difficult to differentiate the boundary \cite{Meijering2012Cell}. In addition, it is easier to collect ground truth data for cell detection compared to segmentation from pathologists who are already under pressure of high work load. In this paper, we present a deep learning approach for cell detection in hematoxylin and eosin (H\&E) stained lung cancer images. 
	
\begin{figure}[t!]
		\centering
		\includegraphics[width=0.6\textwidth]{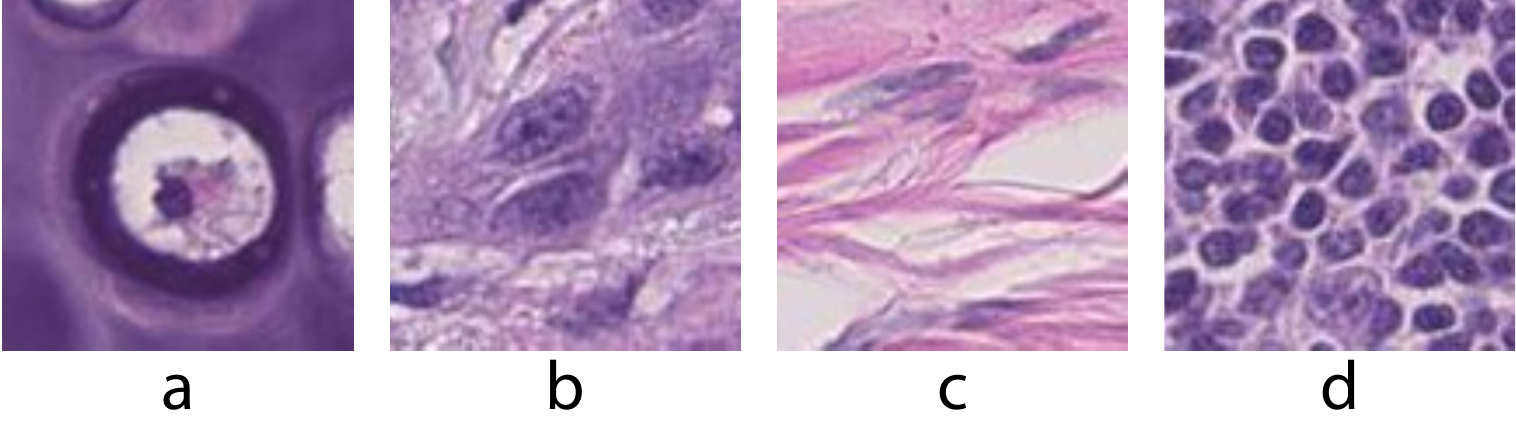}
		\caption{Sample images for various cell types at $20\times$. a. cartilage b. tumour c. stromal d. lymphocytes.} 
		\label{fig:cell_size_sample}
	\end{figure}

	The ground truth obtained for cell detection is usually in the form of dot annotations where each dot represents the centre of a cell. It is easier to solve cell detection as a regression problem \cite{kainz2015you} rather than binary classification of individual pixels which involve complex voting/encoding mechanisms \cite{Xie2015deep,Xue2017}. Therefore, most of the deep learning approaches regress the probability of belonging to the centre of cell followed by local maxima detection \cite{Sirinukunwattana2016,Xie2018Microscopy,Xie2018Efficient}. Local maxima detection is comparatively easier in problems like mitotis detection where the mitotic cells are relatively far apart and probability maps are comparatively sharp \cite{Ciresan2013}. However, when detecting cells across the tumour landscape, the probability maps are not very sharp especially for tumour cells with broken chromatin architecture where there can be multiple local maxima in the probability map. This is similar with other cell types such as fibroblasts and cartilage where width of hematoxylin channel changes through the nucleus. In these cases, a grouping distance can be defined within which there cannot be multiple detections. This parameter is difficult to tune due to large variability in the size of various cell types. Figure \ref{fig:cell_size_sample} illustrates different cell types in cases where there is no optimal cell size to choose for grouping distance parameter.
	
	In this paper, we propose an alternative approach to solve the above discussed problem. We 1. generate artificial mapped labels by convolving dot annotations with a mapping filter. 2. train a CNN for mapped labels and then 3. deconvolve the output of the trained CNN with the same mapping filter to retreive cell coordinates. The results show that the proposed method performs better in comparison with other state-of-the-art deep learning methods in terms of precision and F1-Score.
	
	\subsection{Related Work}
	Traditional hand crafted feature based approches rely on morphological features such as thresholding, region growing, level sets, clustering and graph cuts \cite{Veta2014}. Cosatto \textit{et al}. \cite{Cosatto2008grading} utilised the difference of Gaussian (DoG) filter for cell detection followed by hough transform to detect the peaks. Al-Kofahi \textit{et al}. \cite{alkofahi2010improved} employed graph-cut based method initialised by seeds extracted from Laplacian of Gaussian (LoG) filter. Kuse \textit{et al}. \cite{Kuse2011local} porposed local isotropic phase symmetry for detection of beta cells in pancreas. Yuan \textit{et al}. \cite{Yuan2012} proposed marker controlled watershed with seeds detected by thresholding. Veta \textit{et al} \cite{Veta2013} use fast radial symmetry transform to identify nuclei centres. Arteta \textit{et al}. \cite{Arteta2012learning} utilised maximally stable extremal regions for detection of nuclei. Ali \textit{et al}. \cite{Ali2012an} proposed active contours for cell detection and segmentation. 
	
	Deep learning methods have become a method of choice due to their promising results when dealing with large data sets \cite{Litjens2017asurvey}. Cireșan \textit{et al}. \cite{Ciresan2013} presented one of the early methods utilising deep learning for mitosis detection in breast cancer images. They trained a CNN to regress probability of each pixel belonging to \textit{mitosis} or \textit{non-mitosis}. Cruz-Roa \textit{et al}. \cite{Cruz-Roa2013} and Xu \textit{et al}. \cite{Xu2016} learn unsupervised features using auto-encoders which are fed to a classifier for cell detection. Wang \textit{et al}. \cite{Wang2014} extended this method by cascading CNN and hand-crafted features for mitosis detection.  Xie \textit{et al}. \cite{Xie2015deep} proposed to localise nuclei centres using a voting mechanism. Sirinukunwattana \textit{et al}. \cite{Sirinukunwattana2016} proposed a spatially constrained CNN (SCCNN) by appending two extra layers to the fully connected layer. The added spatially constrained layers estimate the probabilty of a pixel being the centre of a nucleus. Kashif \textit{et al}. \cite{Kashif2016} extended this framework by adding hand-crafted features to the input which slightly improved the F1-score and recall at the expense of precision. Chen \textit{et al}. \cite{Chen2016} proposed a deep regression network which learns it's parameters for a promixity map generated by the segmentation mask of mitotic cells. Wang  \textit{et al}. \cite{Wang2016} proposed a combination of two CNNs which perform simultaneous detection and classification of cells. Xie \textit{et al}. \cite{Xie2018Microscopy} regress a cell density map followed by local maxima detection to detect cells. Recently, Xue \textit{et al}. \cite{Xue2017} proposed a CNN which regresses an encoded feature vector that can be used to recover sparse cell locations. These detections are combined to get the final detection point. Xie \textit{et al}. \cite{Xie2018Efficient} proposed structured regression to learn proximity maps with higher values near cell centres, the local maxima provides the centre of cell location.

	\section{The Proposed Method}
\begin{figure}[t!]
		\includegraphics[width=0.85\textwidth]{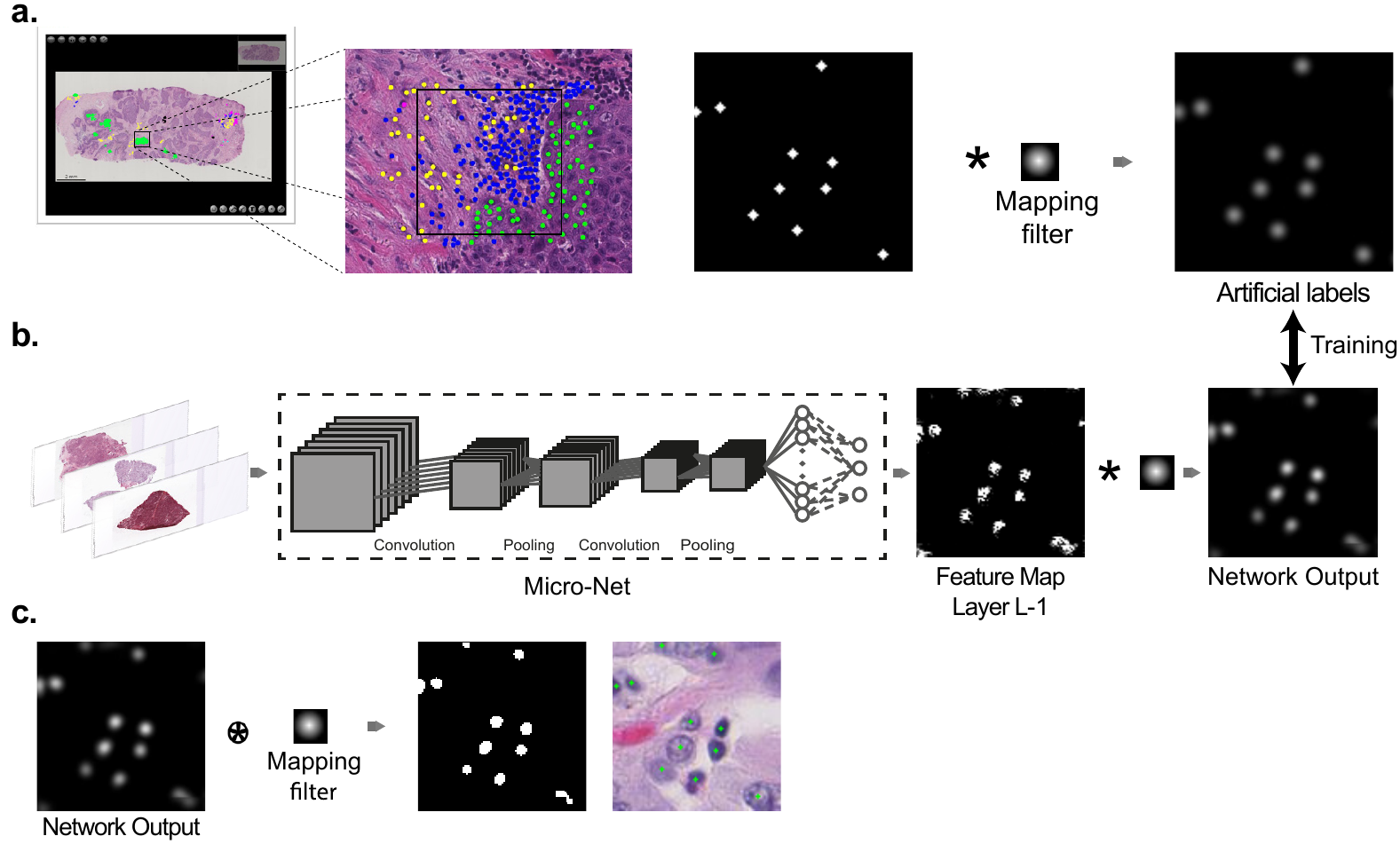}
		\caption{Overview of the proposed method. a. An expert pathologist annotates various cell types using a web based interface which is translated to dot annotations for cell detection. The binary dot labels are convolved with a mapping filter to generate artificial mapped labels.  b. The output layer of Micro-Net \cite{Raza2018} is convolved with the same mapping filter and the modified network is trained for artificial labels from (a). c. The output of the trained network is deconvolved and thresholded with the mapping filter to obtain a binary mask, the centroids of the mask represent the centre of a cell.} 
		\label{fig:Proposed_approach}
	\end{figure}

	The overview of proposed method is shown in Figure \ref{fig:Proposed_approach} and can be divided into three parts: a) Map dot annotations, b) Train the network for mapped labels, c) Deconvolve output of trained network to obtain cell detections.
	
	\subsection{Map dot annotations}
	To create a mapping filter, we observed that the average diameter of smallest abundant cell (lymphocyte) in our data set is about $10$ pixels ($20\times$). We created a binary image $b$ of size $11 \times 11$ with centre of the  image at location ($6$, $6$) set to $1$ and the rest of pixels to $0$. We defined radius $r=5$ pixels to generate the mapping filter $f$ 
	\begin{equation}
	\label{eq:mapping_filter}
	f=\begin{cases}(r - dist(b))/r & dist(b)  \geq  r\\0 & otherwise\end{cases} 	
	\end{equation}
	where $dist(b)$ defines the Euclidean distance transform of the binary image. The resulting mapping filter is similar to a probability map with the maximum value ($1$) at the centre of the image, while the probability reduces as we move away from the centre. As a dot can be recognised as a point source in a binary image, convolving dot annotations with the mapping filter $f$ have a similar effect as the point spread function (PSF) of a lens when light passes through it.
	
	\subsection{Network Training}
	To obtain the probability map of the same size as the input image we perform pixel-wise regression using Micro-Net \cite{Raza2018} ($252 \times 252$) architecture which has recently been shown to be efficient compared to U-Net \cite{Ronneberger2015} and other state of the art pixel-wise classification approaches. Another reason for choosing this architecture is its ability to visualise the input at multiple resolutions which is necessary to train for various cell sizes. In addition, inspiration from U-Net architecture in its design incorporates context information during training from neighbouring cells which is missed by patch-based algorithms such as SCCNN. We modify Micro-Net by adding an additional layer to the output $L-1$ which convolves $L-1$ with the mapping filter as shown in Figure \ref{fig:Proposed_approach}(b), where $L$ is the number of layers in the modified network. This mapping filter is fixed during training which helps to force the output to match the shape of artificial labels and the feature map at $L-1$ towards binary dot annotations. In addition to the mapping filter, we use rectified linear unit (RELU) activation instead of $tanh$ as we need positive values at $L-1$. The modified network is then trained for the mapped labels using Adagrad optimisation and weighted cross entropy loss function where the positive weight was empirically chosen to be 100. \cite{tensorflow2015-whitepaper,duchi2011adaptive,Raza2018}. 
	
	\subsection{Deconvolving the Network Output}
	\label{sec:dconv}
	Once trained, the output of the network can be considered as a ``blurred" image of a point source (centre of a cell) with point spread function (PSF) defined in Equation \ref{eq:mapping_filter}. The output is refined by normalisation by the maximum value and then deconvolved with blind deconvolution algorithm \cite{hanisch1996deconvolution} with an inital PSF provided by the mapping filter. The output is thresholded to obtain binary regions. The centroid of each region is obtained as the centre of a detected cell (Figure \ref{fig:Proposed_approach}(c)). 
	
	\section{Experiments and Results}
	The proposed method was implemented in tensorflow version 1.8 \cite{tensorflow2015-whitepaper}. The data set used in this paper is extracted from histology samples of lung cancer patients from the TRACERx study \cite{Mariam2014Tracking}. Sample images from 55 whole slide images from different patients were extracted to incorporate stain variation and represent various cell types according to their populations.  We used $13,484$ ($\sim$70\%) annotations from $99$ sub-images for training and $5,672$ ($\sim$30\%) annotations from $26$ sub-images for testing. Minimum width and height of each sub-image was set to $252$ with no limit on maximum size. The learning rate was empiricially set to $0.001$ and the network was trained for $200$ epochs. Data augmentation was performed using random crop, left/right and up/down flip. In addition we added random variations for brightness, contrast, hue and saturation using tensorflow implementation.
	
\begin{figure}[t!]
		\includegraphics[width=1.0\textwidth]{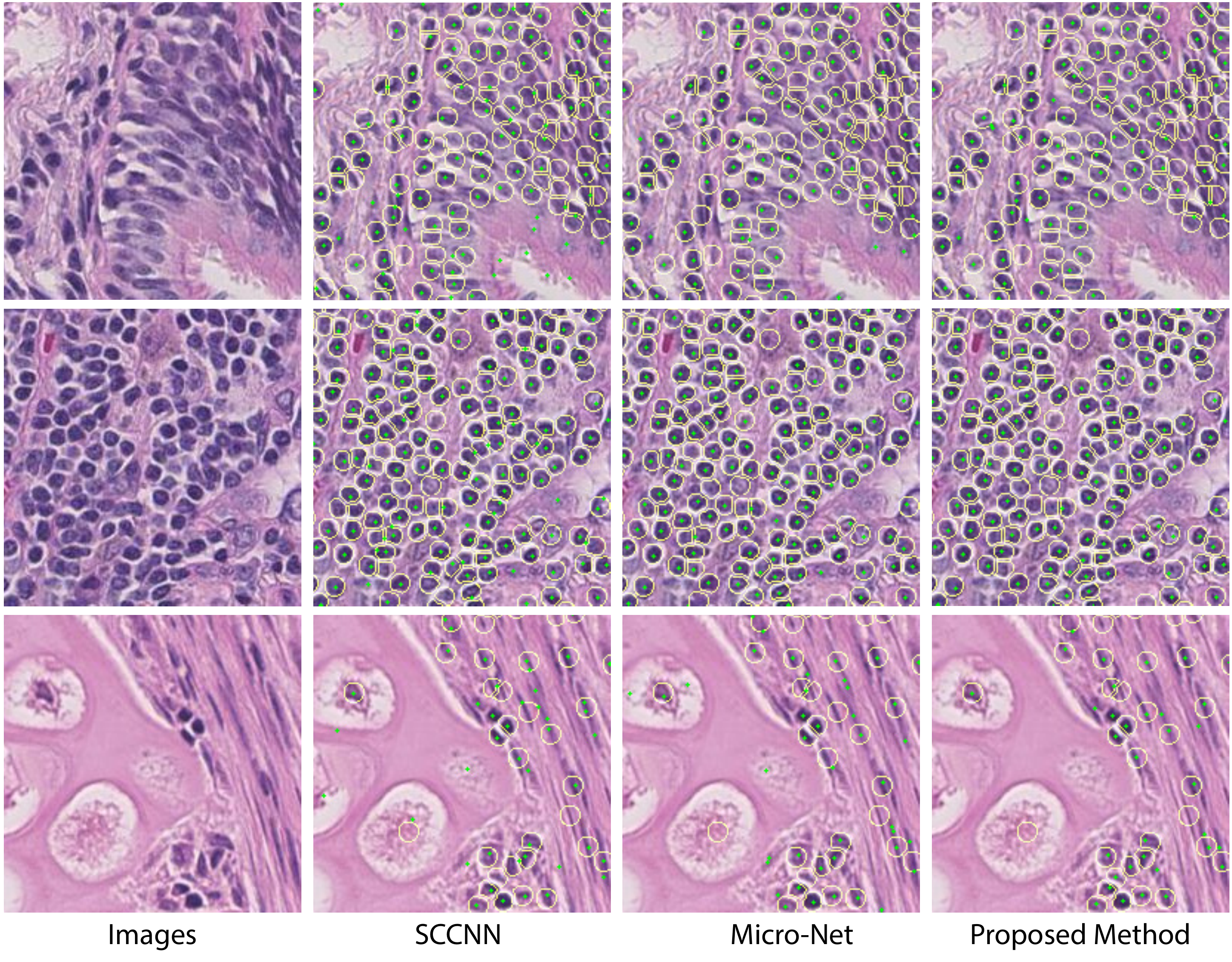}
		\caption{Comparison of proposed method with SCCNN \cite{Sirinukunwattana2016} and Micro-Net \cite{Raza2018} for cell detection. Yellow boundary outlines $6$ pixel radius around the ground truth annotation and the green dots represent output. The first row represents a tumour region, second lymphoid structure and third represents a challenging example of cartilage and stromal region. } 
		\label{fig:qualitative_results}
	\end{figure}

	We compare our method with the recently proposed SCCNN \cite{Sirinukunwattana2016} and Micro-Net networks \cite{Raza2018}. Micro-Net is a segmentation algorithm but the centroid of each segmented cell can be used as a detection point. Qualitative results for the proposed method are shown in Figure \ref{fig:qualitative_results}  where yellow circles represent $6$ pixel radius \cite{Sirinukunwattana2016} around the ground truth and the green dots represents the output of respective algorithm. In the first row, SCCNN produces false positives in the bottom right corner but misses a few cells in the middle where tumour cells lose the chromatin texture. This is due to its higher sensitivity to hematoxylin channel. In addition, it detects the location of a cell at merging cell boundaries indicating that it may not be able to differentiate between two local maxima. This problem occurs when SCCNN is not able to incorporate context information while training and the peaks might not be very sharp. Micro-Net also misses a few cells in the middle and top right of the image. In the second row, SCCNN shows similar kind of problem where it detects cells at boundary of closely packed lymphocytic cells. Compared to the proposed method, Micro-Net performs slightly better in this case where the proposed method misses a few nuclei. The third image is a challenging example consisting of large cartilage cells and spindle like stromal nuclei. SCCNN and Micro-Net both struggle in this case producing quite a few false positives. In summary, SCCNN is highly sensitive to hematoxylin channel, whereas it fails to detect the centre of a cell where the nuclei are closely placed. Micro-Net produces better results with round/oval shaped nuclei but struggles with irregular shaped nuclei specially spindle shaped stromal cells. The proposed method although misses a few nuclei in regions with closely packed cells but overall it performs better with far less false positives.
	
	For a quantitative comparison all detected points which fall within the $6$ pixel radius \cite{Sirinukunwattana2016} around the ground truth dot annotations (i.e., within the yellow boundary in Figure \ref{fig:qualitative_results}) are considered as true positives. The quantitative results are shown in table \ref{tab:quantitative_comparison}. F1-score for SCCNN was calculated to be $78.10$ ($80.2\%$ in \cite{Sirinukunwattana2016}) and for Micro-Net about $83.23\%$, both methods' score significantly lower as compared the proposed network. For quantitative comparison, we used both local maxima \cite{Xie2018Efficient} and the deconvolution to extract the true detection points from the proposed network. Both approaches produced an F1-score of about $\sim86.7\%$. However, local maxima has lower precision and higher recall compared to the deconvolution approach. This means that the latter produces less false positives ($608$) compared to the former ($780$) which is often desired. This comes at the expense of higher false negatives with the second approach ($866$ compared to $733$), hence lower recall. 
	
	\begin{table}[tbh]
		\centering
		\caption{Quantitave comparison with recent deep learning algorithms.}
		\label{tab:quantitative_comparison}
		\begin{tabular}{lccc}
			\hline
			Method                & Precision        & Recall           & F1-Score         \\ \hline
			SCCNN \cite{Sirinukunwattana2016}                 & 72.00\%          & 85.31\%          & 78.10\%          \\ \hline
			Micro-Net \cite{Raza2018}             & 81.32\%          & 85.24\%          & 83.23\%          \\ \hline
			Proposed (Local Maxima) & 86.36\%          & 87.08\%          & 86.72\%          \\ \hline
			Proposed Method       & \textbf{88.77\%} & \textbf{84.73\%} & \textbf{86.70\%} \\ \hline
		\end{tabular}
	\end{table}
	
	\section{Conclusions}
	We presented a deep learning based method for cell detection in H\&E stained cancer images. The proposed method tackles cell detection as a regression problem, generates artificial labels using a mapping filter, optimises the CNN to train for the artificial labels and then deconvolves the mapped output of the network to detect cells. We compare our method with two recently proposed state-of-the-art methods as well as the conventional local maxima approach. The detection of the deconvolution method results in higher precision and lower number of false positives. In addition, the proposed method does not require preprocessing the images for stain normalisation or deconvolution, thus improving the efficiency. In future, we plan to extend this work for multi-class cell classification.
	
	\section{Acknowledgements}
	The authors would like to thank the TRACERx consortium for sharing the data set \cite{Mariam2014Tracking} for this study. We would also like to thank Prof. Nasir Rajpoot at TIA lab, University of Warwick for their help in implementation of SCCNN \cite{Sirinukunwattana2016}. 
	%
	\bibliographystyle{splncs04}
	\bibliography{mybib}
\end{document}